%% file: main.tex
\documentclass[10pt,twocolumn,letterpaper]{article}

\usepackage{cvpr}              
\usepackage[accsupp]{axessibility} 

\usepackage{multirow}
\input{preamble}

\definecolor{cvprblue}{rgb}{0.21,0.49,0.74}
\usepackage[pagebackref,breaklinks,colorlinks,citecolor=cvprblue]{hyperref}

\def\ours{\texttt{\textbf{VideoSAGE}}}

\def\paperID{4} 
\def\confName{CVPR}
\def\confYear{2024}

\title{VideoSAGE: Video Summarization with Graph Representation Learning}

\author{Jose M. Rojas Chaves\\
Intel Corporation\\
{\tt\small jose.rojas.chaves@intel.com}
\and
Subarna Tripathi\\
Intel Labs\\
{\tt\small subarna.tripathi@intel.com}
}

\begin{document}
\maketitle
\input{sec/0_abstract}    
\input{sec/1_intro}

\input{sec/2_related}
\input{sec/3_method}
\input{sec/4_results}
\input{sec/5_conclusion}

{
    \small
    \bibliographystyle{ieeenat_fullname}
    \bibliography{main}
}


\end{document}

%% file: preamble.tex
\usepackage[dvipsnames]{xcolor}


%% file: sec/0_abstract.tex
\begin{abstract}

We propose a graph-based representation learning framework for 
video summarization. First, we convert an input video to a graph where nodes correspond to each of the video frames. Then, we impose sparsity on the graph by connecting only those pairs of nodes that are within a specified temporal distance. 
We then formulate the video summarization task as a binary node classification problem, precisely classifying video frames whether they should belong to the output summary video. A graph constructed this way aims to capture long-range interactions among video frames, and the sparsity ensures the model trains without hitting the memory and compute bottleneck. 
Experiments on two datasets(SumMe and TVSum) demonstrate the effectiveness of the
proposed nimble model compared to existing state-of-the-art 
summarization approaches while being one order of magnitude more efficient in compute time and 
memory.  
 
\end{abstract}

%% file: sec/1_intro.tex
\section{Introduction}
\label{sec:intro}

The landscape of video creation and consumption has been drastically changed in the last decade thanks to the affordable video capturing devices and the wide spread use of the Internet. Recent years have seen significant surge of social networks and video streaming, causing prevalence of user-created videos. Quick video browsing among massive video contents thus become essential. Video summarization is a way to facilitate quick grasping of video content by squeezing the most salient content from a long video to a short one.  

\begin{figure}[ht]
    \centering
    \includegraphics[width=1\linewidth]{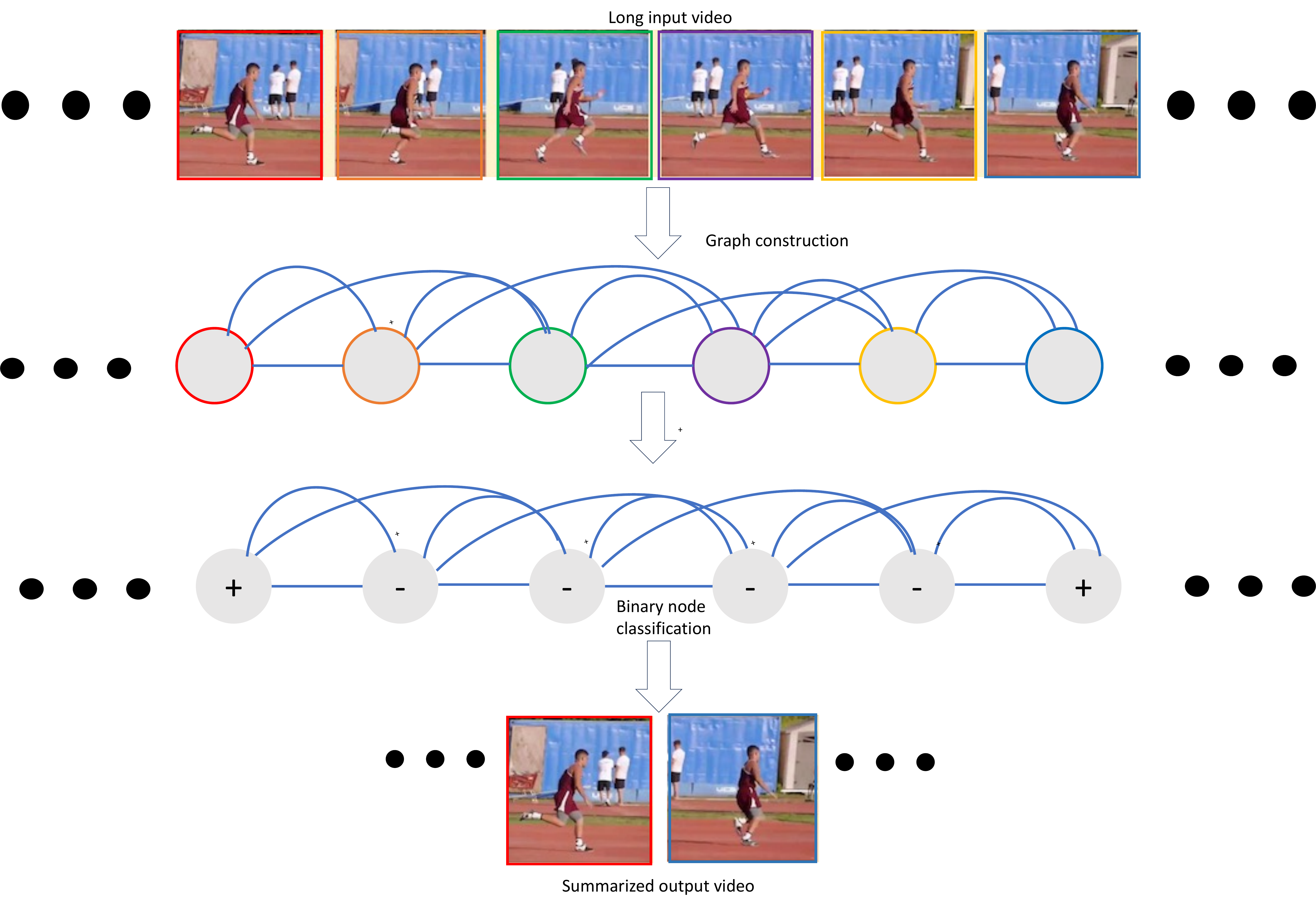}
    \caption{ {\ours} constructs a graph from the input video, where each node corresponds to a video frame. Only those pairs of nodes are connected to each other who are within a temporal distance. 
    Video summarization is thus formulated as a binary node classification problem for that graph. Our constructed graph has forward, backward and bidirectional edges. For visual clarity, we only show bi-directional edges in this figure. From top row to bottom row, the figure shows how regular input video is converted to a sparse graph, followed by binary node classification on nodes leading to summarized output video.}
    \label{fig:framework}
\end{figure}

To retain the most informative information of a video to its summarized version, we propose a framework leveraging short-range and long-range correlation among video segments. Our framework takes an input video and internally converts it to a graph. We impose sparsity constraints on the graph, so the internal representations remains manageable in a standard compute system, as well as retain its power of expressivity while learning from short-range and long-range correlations among different temporal segments. To this end, we propose {\ours}, where each of the video frames become nodes of a graph and a subset of node pairs are connected to each other. We then learn parameters for graph convolutional networks on the internal graph representations and optimize our model for binary node classification signifying informative vs non-informative nodes for generating the video summary. Figure \ref{fig:framework} demonstrates the framework pictorially. 
Please note that there is no ground truth graph. We can only compare our summarization model based on the end performance measured by summarization and correlation metrics, and can not objectively assess the graph construction otherwise. 

We perform experiments on two popular summarization datasets, namely TVSum and SumMe, and show the efficacy of our proposed method by the objective scores such as correlation metrics and F-1 scores. We also demonstrate the qualitative results of the generated summarized videos. Most importantly, comparing with the existing state-of-the-art methods, {\ours}  provides an order of magnitude faster inference time, smaller model size; 3$\times$ savings in peak memory footprint.  

Our implementation is based on an open sourced library and the source code of our method is available on GitHub: \url{https://github.com/IntelLabs/GraVi-T}. 

The novelty of our approach is in formulating the video summarization problem as a node classification on a graph. We construct the graph such that it enables interactions only between relevant nodes over time. The graph remains sparse enough such that the long-range context aggregation can be accommodated within a comparatively smaller memory and computation budget.

%% file: sec/2_related.tex
\section{Related Work}
\label{sec:related_work}

\textbf{Video summarization} \\
Various machine-learning techniques have been employed for video summarization. These approaches can be categorized into two main groups: supervised~\cite{mavs2018, hsarnn2018, vasnet2019, pglsum2021, msva2021, clipit_2021, dsnet2021, iPTNet, rrstg2022, a2summ2023, sumgraph2020, rsgn2021} and unsupervised~\cite{csnet2018, drdsn2018, rsgn2021, casum2022} methods. 
We primarily focus on the supervised learning methods
in this section. \\
Numerous models in supervised learning have utilized annotated video datasets such as TVSum~\cite{TVSum} and SumMe~\cite{SumMe} for video summarization. One notable example of these models is A2Summ~\cite{a2summ2023}, which proposed a multimodal transformer-based model that aligns and attends multiple inputs (e.g., video, text, and sound) to select their most important parts. PGL-SUM~\cite{pglsum2021} combined global and local multi-head attention mechanisms with a regressor network to select keyframes. CLIP-it~\cite{clipit_2021} used a language-guided multi-head attention mechanism in addition to a multimodal transformer to generate summaries based on natural language queries. iPTNet~\cite{iPTNet} proposed an importance propagation based Teaching Network consisting of two separate collaborative modules that conducted video summarization and moment localization. In general, many other supervised learning methods rely also on self-attention mechanisms~\cite{mavs2018, vasnet2019, msva2021, dsnet2021} or LSTM networks~\cite{mavs2018, hsarnn2018, dsnet2021} for frame importance prediction. Most state-of-the-art (SOTA) methods either use multimodality or a combination of multiple neural network techniques, making these methods relatively more complex. On the other hand, our proposed method achieves comparable results with a only fraction of memory and compute cost. \\

\noindent
\textbf{Graph-based video representation} \\

While transformer models have recently took center stage in the research field of video understanding, in recent times, graph neural networks operated on explicit graph based representation are emerging for their complementary traits, including long-form reasoning thanks to the inherent low memory and compute requirements. 
Applications of specialized GNN based models in video understanding includes:visual relationship forecasting~\cite{mi2021visual}, dialog modelling~\cite{geng2020spatio}, video retrieval~\cite{tan2021logan}, emotion recognition~\cite{shirian2020learnable}, action detection~\cite{zhang2019structured}, video summarization~\cite{sumgraph2020}, and others~\cite{nagarajan2020ego,patrick2021space,min2022intel,min2023sthg}.

Only a few methods utilize Graph Neural Networks (GNNs) for video summarization. RR-STG~\cite{rrstg2022} built spatial and temporal graphs over which it performed relational reasoning with graph convolutional networks (GCNs) and extracted spatial-temporal representations for importance score prediction and key shot selection. RSGN~\cite{rsgn2021} contained a summary generator and a video reconstructor. The first layer of the generator is a bidirectional LSTM that encodes the frame sequence in each shot, and the second layer is a graph model to explore the dependencies among different shots. And SumGraph~\cite{sumgraph2020}proposed a recursive graph modeling network consisting of 3 GCN layers plus another GCN working as a summarization layer. Unfortunately, no source code was found for any of these graph-based methods for results replication.\\
With only some exceptions~\cite{hsarnn2018, sumgraph2020, msva2021, rsgn2021}, most of these methods concentrated on keyframe predictions (including our proposal), and simply relied on using the knapsack algorithm, over predefined segments built with kernel temporal segmentation (KTS) and its respective predicted importance scores, for creating the final video summaries. 
The above context will be relevant while discussing the evaluation methods in section~\ref{subsec:evaluation_metrics}.

Our approach {\ours } is distinguished from the literature in the way of formulating the video summarization problem as a node classification on a graph constructed from the input videos. 
Our constructed graphs are  sparse enough such that the long-range context aggregation can be accommodated within a comparatively smaller memory and computation budget, while capable of modeling short-range and long-range context aggregation. 

%% file: sec/3_method.tex
\section{Methodology}

\begin{figure*}[ht]
    \centering
    \includegraphics[width=\textwidth]{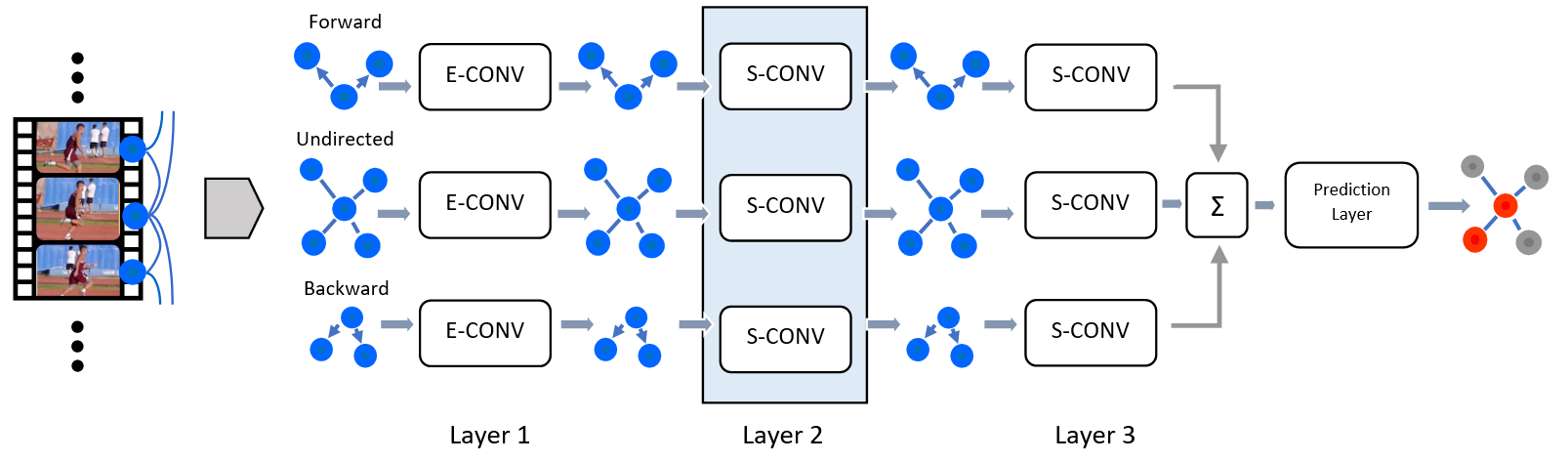}
    \caption{An illustration of utilized \textit{Bi-directional} (a.k.a. \textit{Bi-dir}) GNN model for video summarization.  Here, we have three separate GNN modules for the forward, backward, and undirected graph, respectively. Each module has three layers where the weight of the second layer is shared by all three graph modules. The second layer is placed inside a solid-lined box to indicate the weight sharing while for the first and the third layer we use dotted-lines. E-CONV and S-CONV are shorthand for EDGECONV and SAGE-CONV, respectively.}
    \label{fig:model}
\end{figure*}

\label{sec:method}

Figure~\ref{fig:framework} illustrates how {\ours} constructs a graph from an input video where each node corresponds to a frame of the video.
This graph is able to reason over long temporal contexts over all nodes inspite of being not fully-connected. This is an important design choice to reduce memory and computation overheads \cite{spell2022,minintel}. The edges in the graph are only between \emph{relevant} nodes needed for message passing, leading to a sparse graph that can be accommodated within a small memory and computation budget. 

We cast the video summarization problem as a node classification problem. To that end, we train a 
lightweight GNN to perform binary node classification on this graph. 
Inspired by  ~\cite{spell2022}, our model utilizes three separate GNN modules for the forward, backward, and undirected graph, respectively. Each module has three layers where the weight of the second layer is shared across all the above three modules.

\subsection{Notations}
\noindent
Let $G=(V,E)$ be a graph with the node set $V$ and edge set $E$. For any $v\in V$, we define $N_v$ to be the set of neighbors of $v$ in $G$. We assume the graph has self-loops, i.e., $v\in N_v$. Let $X$ denote the set of given node features $\{\mathbf{x}_v \}_{v\in V}$ where $\mathbf{x}_v\in \mathbb{R}^d$ is the feature vector associated with the node $v$. 
We can now define a $k$-layer GNN as a set of functions $\mathcal{F}=\{f_i\}_{i\in [k]}$ for $i\geq1$ where each $f_i:
V\rightarrow \mathbb{R}^m $ ( $m$ will depend on layer index $i$). 
All $f_i$  is parameterized by a set of learnable parameters. Furthermore, $X^{i}_V=\{\mathbf{x}_v\}_{v\in V}$ is the set of features at layer $i$ where $\mathbf{x}_v=f_i(v)$. Here, we assume that $f_i$ has access to the graph $G$ and the feature set from the last layer $X^{i-1}_V$. 
\begin{itemize}
    \item $\sf SAGE\text{-}CONV$ aggregation:
    This aggregation\cite{hamilton2018inductive} is one of the widely used GCN type and has a computationally efficient form. Given a $d$-dimensional feature set $X^{i-1}_V$
    , the function $f_i:V \rightarrow \mathbb{R}^m$ is defined for $i\geq1$ as follows:
    $$f(v)=\sigma \Big(\sum_{w\in  N_v}{\sf M}_i\mathbf{x}_w\Big)$$
 where $\mathbf{x}_w\in X^{i-1}_V$, ${\sf M}_i \in \mathbb{R}^{m\times d}$ is a learnable linear transformation, and $\sigma:\mathbb{R}\rightarrow\mathbb{R}$ is a non-linear activation function applied point-wise. 
    
    \item $\sf EDGE\text{-}CONV$ aggregation:
    $\sf EDGE\text{-}CONV$~\cite{edgeconv2019} models global and local-structures by 
    applying channel-wise symmetric aggregation operations on the edge features associated with all the outgoing edges of each node.
    The aggregation function $f_i:V\rightarrow \mathbb{R}^{m}$ can be defined as:
    $$f_i(v)=\sigma \Big(\sum_{w\in  N_v}{\sf g}_i\big( \mathbf{x}_v\circ \mathbf{x}_w
 \big)	\Big)$$ where $\circ$ denotes concatenation and ${\sf g}_i: \mathbb{R}^{2d}\rightarrow \mathbb{R}^m $ is a 
learnable transformation. Usually, ${\sf g}_i$ is implemented by Multilayer perceptrons (MLPs). 
The number of parameters for $\sf EDGE\text{-}CONV$ is larger than $\sf SAGE\text{-}CONV$. This gives the $\sf EDGE\text{-}CONV$ layer more expressive power at a cost of higher complexity.
For our model, we set ${\sf g}_i$ to be an MLP with two layers of linear transformation and a non-linearity. 
    
\end{itemize}

\subsection{Video as a graph} \label{sec:vgraph}
\noindent
We represent a video as a graph that is suitable for the task of summarization. In our implementation, the entire video is represented by a single graph \textit{i.e.}if the video has $n$ frames in it, the graph will have $n$ nodes. We construct one graph for each video in the set.

The node set of $G=(V,E)$ is $V=[n]$, and for any $(i,j)\in [n]\times [n]$, we have $(i,j)\in E$ if the following two condition is satisfied: $|{\sf Time}(i)$-${\sf Time}(j)| \leq T$ where $\sf Time$ is the time-stamp of each video frame and $T$ is a hyperparameter for the maximum time difference between any connected node pairs. 
In other words, we connect two nodes (video frames) if they are temporally near by. Thus, the interactions between different video frames beyond just consecutive frames can be be modeled. 
To pose the video summarization task as a node classification problem, we also need to specify the feature vectors for each node $v\in V$.
As done in several other SOTA methods such as ~\cite{SumMe, pglsum2021}, we also use GoogLeNet ~\cite{GoogleNet} as the visual features for each video frame. 
By notation, a feature vector of node $v$ is defined to be $x_v=[v_{{\sf visual}}]$ where $v_{{\sf visual}}$ is the GoogLeNet feature of the video frame $v$. Finally, we can write $G=(V,E,X)$ where $X$ is the set of the node features.

\subsection{Video summarization as a node classification task} \label{subsec:task}
\noindent
In the previous sub-section, we have described our graph construction procedure that converts a video into a graph $G=(V,E,X)$ where each node has its own visual feature vector. During the training process, we have access to the ground-truth labels of all video frames  indicating whether they belong to the summarized output or not. Therefore, the task of video summarization can be naturally posed as a binary node classification problem in the above mentioned graph $G$, whether a node belongs to the output summary or not. Specifically, we train a three-layer GNN for this classification task (See Figure~\ref{fig:model}). The first layer in the network uses ${\sf EDGE\text{-}CONV}$ aggregation to learn pair-wise interactions between the nodes. For the last two layers, we observe that using ${\sf SAGE\text{-}CONV}$ aggregation provides better performance than ${\sf EDGE\text{-}CONV}$, possibly due to ${\sf EDGE\text{-}CONV}$'s tendency to overfit.

Using the criterion: $|{\sf Time}(i)-{\sf Time}(j)| \leq T$ for connecting the nodes, the resultant graph renders undirected. 
In order to incorporate additional temporal ordering, we explicitly incorporate temporal direction as specified in ~\cite{spell2022}. The undirected GNN is augmented with two other parallel networks; one for going forward in time and another for going backward in time.

Specifically, in addition to the undirected graph, we create a forward graph where we connect $(i,j)$ if and only if $0\geq {\sf Time}(i)-{\sf Time}(j) \geq- T$. Similarly, $(i,j)$ is connected in a backward graph if and only if $0\leq {\sf Time}(i)-{\sf Time}(j) \leq T$. This gives us three separate graphs where each of the graphs can model different temporal relationships between the nodes. 
Additionally, the weights of the second layer of each graph is shared across the three graphs. This weight sharing technique can enforce the temporal consistencies among the different information modeled by the three graphs as well as reduce the number of parameters. 

%% file: sec/4_results.tex

\section{Experiments and Evaluation}
\label{sec:results}

\subsection{Experimental set up}
\label{subsec:experimental_setup}

\textbf{Datasets} \\
We use two bench-marking datasets to evaluate the performance of our proposed {\ours} model. A copy of these datasets are downloaded from PGL-SUM’s repository\footnote{https://github.com/e-apostolidis/PGL-SUM/tree/master/data}. The SumMe~\cite{SumMe} dataset is composed of 25 raw videos (1 to 6.5 minutes duration) covering holidays, events, and sports. And the TVSum~\cite{TVSum} dataset consists of 50 YouTube videos (2 to 10 minutes duration) covering 10 categories selected from the TRECVid Multimedia Event Detection dataset ~\cite{potapov2014category}. Both datasets were originally evaluated by 15-20 different human users. And video summaries were built from each user evaluation by using the knapsack algorithm. Additionally, every video was down-sampled to 2 fps and provided with sampled-level importance scores (averaged from all users’ inputs) and features (Size D = 1024) extracted from GooLeNet’s~\cite{googlenet2015} Pool 5 layer. \\\\
\textbf{Experiments}\\
To prepare the data inputs for our experiments, we create graph representations from each video in the datasets where each node corresponds to a frame within a temporal window of that video. This is done using \href{https://github.com/IntelLabs/GraVi-T}{\textbf{GraVi-T}}\footnote{https://github.com/IntelLabs/GraVi-T} so the graph can reason over long temporal contexts for all nodes without being fully connected. Edges in the graphs are formed between temporally nearby nodes, up to 10 adjacent (backward and forward) frames  for TVSum and 20 frames for SumMe, creating a sparse graph. \\
Each dataset was arranged into 5 splits following PGL-SUM’s~\cite{pglsum2021} approach. For each split, 20\% of the videos were randomly selected for the validation set with the remaining 80\% being used for training. 
Then, we train a lightweight GNN to perform binary node classification 
on each video graph, running 40 (SumMe) or 50 (TVSum) epochs over one split, the final model for one split is taken from the epoch with the lowest validation loss. Finally, evaluation metrics are gathered by running that model onto its validation set. This process is repeated over each split and the final evaluation results are the averages from all the splits. \\
To find a good set of hyper-parameters we follow an iterative process and utilize on RayTune’s~\cite{raytune2018} grid search tool for exploration. The focus of the process is on finding stable learning curves and maximizing the Kendall's $\tau$ correlation result. The selected learning rate and weight decay values for SumMe were 0.001 and 0.003 respectively. For TVSum they were 0.002 and 0.0001 respectively.  Batch size and dropout rate for both cases were 1 and 0.5 respectively. 

\subsection{Evaluation Metrics}
\label{subsec:evaluation_metrics}

\begin{table}[t!]
    \caption{Comparison of results from uniformly randomly generated importance scores and results from perfect predictions (ground truth) using three evaluation methods on TVSum~\cite{TVSum}: F1-Score, Otani et al~\cite{otani2018}, and our proposed method. * denotes reproduced results.}
    
    \label{table:eval_method_table}
    \centering
    \setlength{\tabcolsep}{7pt}
    \resizebox{0.9\linewidth}{!}{
        \begin{tabular}{l|c|cc|cc}
        \toprule
        Predictions & F1-Score & \multicolumn{2}{c|} {Otani et al~\cite{otani2018}*} & \multicolumn{2}{c} {Proposed Method} \\ \midrule
        & F1 ($\uparrow$) & $\tau$ ($\uparrow$) & $\rho$($\uparrow$) & $\tau$($\uparrow$) & $\rho$($\uparrow$) \\ \midrule
          Random  & $54.37$   &   $0.00$ & $0.00$   &   $-0.006$ & $-0.009$ \\
          Perfect & $62.87$   &   $0.37$ & $0.46$   &   $ 1.000$ & $ 1.000$ \\ 
          \bottomrule
        \end{tabular}
    }
\end{table}

Most existing SOTA methods have used the F1-Score and sometimes correlation metrics as well, to evaluate their models. The F1-Score is calculated as the maximum (SumMe) or the average (TVSum) value by comparing the predicted summary against each provided user summary from a given video, and then averaged from all videos in the validation set (as in ~\cite{pglsum2021, a2summ2023}). As pointed out in section 2, in most cases these summaries are created by running the Knapsack algorithm over predefined segments obtained through KTS. 
However, as stated by Otani \textit{et al.}~\cite{otani2018}, the F1-Score result is mainly dictated by the video segmentation and its segment lengths, resulting into the contribution of the importance scores 
being completely ignored by the benchmark tests. This means that many SOTA works have been evaluating and comparing their models with a method that give more weight to that step they all have in common (Knapsack) and that can even ignore the piece they focus most and try to differentiate for the model.\\
Otani et al~\cite{otani2018} proposes Spearman’s $\rho$ and Kendall’s $\tau$ as correlation coefficients to evaluate models on how close the predicted scores are to human annotated scores. The correlation score for one video is then obtained by averaging the results over each individual annotations (20 user annotations on TVSum) and then the final score is obtained by averaging over the validation set. \\
An argument provided by Otani et al~\cite{otani2018} against using the F1-Score method is that even using uniformly randomly generated predictions would result in a relatively high score (54.37\%), comparable to other models at the time. On the other hand, calculating the respective correlation coefficients with this method would result in a near zero value, as one would have expected. 
See Table~\ref{table:eval_method_table}. 
However, if the model were perfect enough to predict its ground truth, Otani’s method won’t result in near perfect result ($\sim$1.0). This is because the result is the average over all user annotations, and it is impossible for the model to predict all of them at the same time. These annotations are subjective and they differ from user to user. In fact, the maximum value a model could technically 
achieve on TVSum is (62.87\%). \\
We agree with Otani et al. about the F1-Score and provide its result values here only for reference. 
We choose a more traditional way of evaluating our model, that is comparing against the ground truth importance scores. Which we consider as a fairer approach since it was what the model was trained to predict.
A perfect prediction with this method would give a perfect result (1.0), as seen in Table~\ref{table:eval_method_table}. Providing a complete [-1,1] range for evaluation and comparison.  

\subsection{Results}
We compare the proposed
method {\ours} with the previous SOTA
methods on SumMe~\cite{SumMe} and TVSum~\cite{TVSum} datasets as shown in Table~\ref{table:main_table}. A2Summ~\cite{a2summ2023} and PGL-Sum~\cite{pglsum2021} are two of the best methods bench-marked on the above two datasets. For a fair comparison, we have replicated results from these two models to calculate correlation coefficients by following the method proposed in section \ref{subsec:evaluation_metrics}.

Results from table~\ref{table:main_table} show that even when  A2Summ~\cite{a2summ2023} and PGL-Sum~\cite{pglsum2021} have better F1-Scores, it is {\ours} which predicts the importance scores better. {\ours } beats both the above models, on both datasets, on their Kendall's $\tau$ and Spearman's $\rho$ correlation coefficients by 3-4\%, showing its superiority. 
In the following section (\ref{subsec:qualitative_results}), we will exemplify why having better F1-Score does not always mean better prediction or better model. 

\begin{table}[t!]
    \caption{Comparison with SOTA methods on the
    SumMe~\cite{SumMe} and TVSum~\cite{TVSum} datasets. We include the reproduced results of
    methods using GoogLeNet~\cite{GoogleNet} features for a fair comparison. 
    * denotes reproduced results. 
    }

    \label{table:main_table}
    \centering
    \setlength{\tabcolsep}{7pt}
    \resizebox{0.9\linewidth}{!}{
    \begin{tabular}{l|ccc|ccc}
        \toprule
        Method & \multicolumn{3}{c|} {SumMe~\cite{SumMe}} & \multicolumn{3}{c} {TVSum~\cite{TVSum}}  \\
        \midrule
        &  $\tau$ ($\uparrow$) & $\rho$($\uparrow$) & F1($\uparrow$) & $\tau$($\uparrow$) & $\rho$($\uparrow$) & F1($\uparrow$) \\
        \midrule
        Random~\cite{otani2018}   &   $0.00$ & $0.00$ & $41.0$         &  $0.00$ & $0.00$ & $57.0$ \\
        \midrule
        A2Summ~\cite{a2summ2023}*  &   $0.09$ & $0.12$ & $55.0$         &  $0.26$ & $0.38$ & $\bf 63.4$ \\ 
        PGL-Sum~\cite{pglsum2021}* &   $0.09$ & $0.12$ & $\bf 55.6$     &  $0.27$ &$0.39$ & $61.0$ \\
        
        \bf{{\ours} (Ours)}         &  $\bf 0.12$ & $\bf 0.16$ & $46.0$  &  $\bf 0.30$ & $\bf 0.42$ & $58.2$ \\
        \bottomrule
    \end{tabular}
    }
\end{table}

\subsection{Qualitative Results}
\label{subsec:qualitative_results}

\begin{figure}[ht]
    \centering
    \includegraphics[width=1\linewidth]{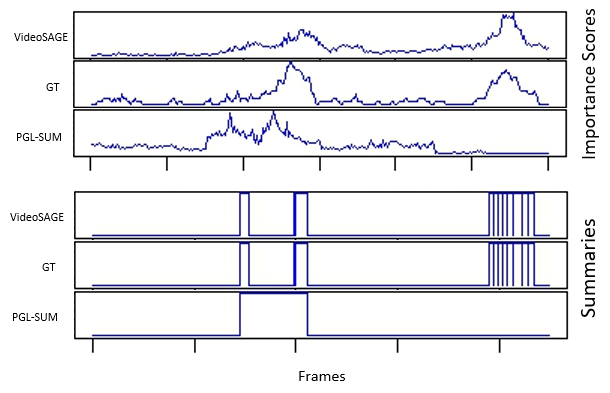}
    \caption{SumMe/\textbf{Video\_1}: Comparison of importance scores and selected summary segments for {\ours} (ours), a ground truth, and PGL-SUM~\cite{pglsum2021}.}
    \label{fig:video1_metrics}
\end{figure}

\begin{figure}[ht]
    \centering
    \includegraphics[width=1\linewidth]{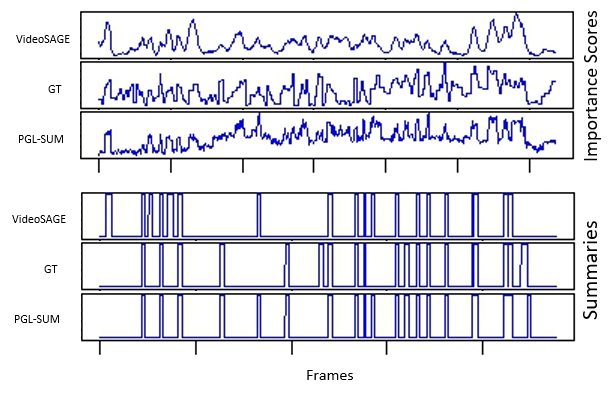}
    \caption{TVSum/\textbf{Video\_16}: Comparison of importance scores and selected summary segments for {\ours} (ours), a ground truth, and PGL-SUM~\cite{pglsum2021}.}
    \label{fig:video16_metrics}
\end{figure}

Two videos were selected to showcase the qualitative results: Video\_16 from TVSum~\cite{TVSum} and video\_1 from SumMe~\cite{SumMe}. Video\_16 was selected as its individual correlation results (Table~\ref{table:eval_videos_table}) where near to the general average (Table~\ref{table:main_table}) on {\ours}. On the other hand, Video\_1 is an outstanding case for {\ours}, showing significantly higher performance on its respective correlation metrics.\\
Following the same reasoning as in section~\ref{subsec:evaluation_metrics}, we are comparing PGL-SUM~\cite{pglsum2021} and {\ours} against ground truth scores. In the case of the predicted summaries, we are comparing against a (GT) summary, generated from the ground truth scores, as opposed to comparing against all user provided summaries. Correlation results for Video\_16 in Table~\ref{table:eval_videos_table} show how PGL-SUM~\cite{pglsum2021} is slightly better than {\ours}. Such results can be qualitatively validated on Figure~\ref{fig:video16_metrics} in which PGL-SUM~\cite{pglsum2021} predicts the GT summary better. In this particular case, better correlation metrics also produced better F1-Score, however, that's not always the case. \\
Video\_1 on Table~\ref{table:eval_videos_table} shows how PGL-SUM~\cite{pglsum2021} have higher F1-Score than {\ours} even though their correlation results are notably lower than those obtained by {\ours}. We see from  Figure~\ref{fig:video1_metrics} how PGL-SUM~\cite{pglsum2021} clearly fails to predict the importance scores on Video\_1 and how the predicted summary does not resemble that of the ground truth scores. Yet, its F1-Score gets benefited from averaging over multiple user summaries. 

\begin{figure}[ht]
    \centering
    \includegraphics[width=1\linewidth]{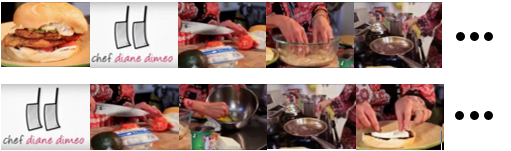}
    \caption{TVSum/\textbf{Video\_16}: Snapshots comparing shots from {\ours} (ours) predicted summary (Top) and a GT summary build from ground truth scores (Bottom).}
    \label{fig:video16_snaps}
\end{figure}

\begin{table}[t!]
    \caption{Comparing results for PGL-SUM~\cite{pglsum2021} and {\ours} on two specific videos from SumMe and TVSum, showing its F1-Score and its Kendall's $\tau$ and Spearman's $\rho$ correlation coefficients. * denotes reproduced results.}
    
    \label{table:eval_videos_table}
    \centering
    \setlength{\tabcolsep}{7pt}
    \resizebox{0.9\linewidth}{!}{
    \begin{tabular}{l|ccc|ccc}
        \toprule
        Model & \multicolumn{3}{c|} {SumMe/\textbf{video\_1}} & \multicolumn{3}{c} {TVSum/\textbf{video\_16}}  \\
        \midrule
        & F1 & $\tau$ & $\rho$ & F1 & $\tau$ & $\rho$ \\
        \midrule
          PGL-SUM*      &  $\bf69.69$ & $0.08$ & $0.10$     &  $\bf68.54$ & $\bf0.32$ & $\bf0.47$ \\ 
          {\ours} (ours)  &  $64.43$ & $\bf0.47$ & $\bf0.60$  &  $62.16$ & $0.31$ & $0.46$ \\
        \bottomrule
    \end{tabular}
    }
\end{table}

\subsection{Parameters Setup} 

\begin{figure}[ht]
    \centering
    \includegraphics[width=1\linewidth]{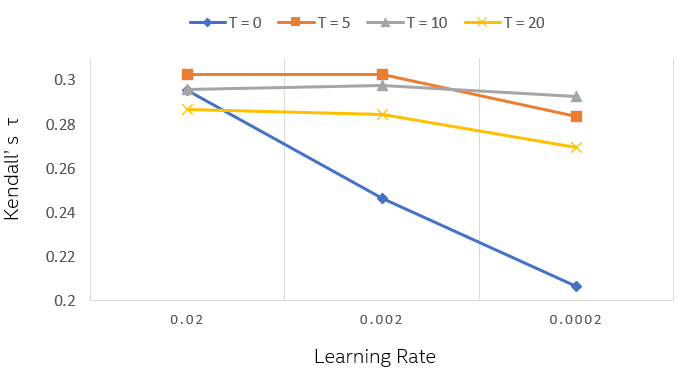}
    \caption{Kendall's $\tau$ correlation results  on TVSum~\cite{TVSum} for different choices of learning rate and number of graph edges (T) per node.}
    \label{fig:hyperparameters}
\end{figure}

As explained in the Experiments section of Section~\ref{subsec:experimental_setup} we utilizeRayTune’s~\cite{raytune2018} grid search to find a proper set of hyper-parameters. Figure~\ref{fig:hyperparameters} shows the effect of different choices of the learning rate (lr) and different T distance values for the number of T forward and T backward connections per node in the graph representation 
on the Kendall's $\tau$ value for videos from TVSum~\cite{TVSum} . These results are the average of 10 repeated experiments per split, totalling to 50 experiment runs per each combination of parameters. \\
The graph representation with best results was achieved for T value of 5. And the best learning rate is between lr=0.02 and lr=0.002. Bigger learning rates than that would result in a very unstable training.
In fact, the standard deviation for T=5 and lr=0.02 was 0.061 while the standard deviation for T=5 and lr=0.002 was 0.051. Kendall's $\tau$ in those two configurations is virtually the same, so the learning rate with more stable results was chosen over the other. 

\subsection{Profiling}
All experiments including PGL-SUM~\cite{pglsum2021}, A2Summ~\cite{a2summ2023} and {\ours} were run on an Intel(R) Core (TM) i9-12900k (3.2 GHz) with 32 GB of memory RAM. The evaluation experiments were run single threaded (for fair comparison) and profiled with PyTorch Profiler. Memory allocation details were extracted by using profiler's \textit{export\_memory\_timeline} command. Results on table~\ref{table:profile_table} are the averages of profiling the inference step during the respective normal evaluation runs of each models. \\
In Table~\ref{table:profile_table}, we can see how {\ours} can do inference an order of magnitude faster than the SOTA while requiring less than 2/5th of the memory allocated by PGL-SUM~\cite{pglsum2021} or A2Summ~\cite{a2summ2023}.

\begin{table}[t!]
    \caption{Comparing profiling results during inference on A2Summ~\cite{a2summ2023}, PGL-SUM~\cite{pglsum2021} and {\ours}. * denotes reproduced results.}
    
    \label{table:profile_table}
    \centering
    \setlength{\tabcolsep}{7pt}
    \resizebox{0.9\linewidth}{!}{
    \begin{tabular}{l|ccc}
        \toprule
         Model &  Average        & Parameters' & Max Memory \\
               &  Inference Time & Memory      & Allocated  \\
               &  (ms)           & (MB)        & (MB)       \\
        \midrule
        PGL-SUM*        &  $113.79$   & $36.02$   & $55.17$    \\ 
        A2Summ*         &  $120.59$   & $9.60$    & $50.56$    \\ 
        {\ours} (ours)  &  $\bf23.55$ & $\bf3.52$ & $\bf19.27$ \\
        \bottomrule
    \end{tabular}
    }
\end{table}

%% file: sec/5_conclusion.tex
\section{Conclusions}
\label{sec:conclusion}

We formulate the video summarization task as a binary node classification problem on graphs constructed from input videos. We first convert an input video to a graph where each node corresponds to a video frame and nodes within a specified temporal distance are connected to each other. We show that this structured sparsity leads to comparable or improved results on video summarization datasets while capable of performing one order of magnitude faster inference with only a fraction of the memory usage comparing with existing methods. \\

\noindent \textbf{Acknowledgement } We thank Kyle Min, who is the developer of \href{https://github.com/IntelLabs/GraVi-T}{\textbf{GraVi-T}}, for his valuable feedback and comments.